%% file: main.tex
\pdfoutput=1
\documentclass[10pt,twocolumn,letterpaper]{article}

\usepackage{cvpr}              

\input{preamble}

\definecolor{cvprblue}{rgb}{0.21,0.49,0.74}
\usepackage[pagebackref,breaklinks,colorlinks,allcolors=cvprblue]{hyperref}

\def\paperID{17835} 
\def\confName{CVPR}
\def\confYear{2025}

\renewcommand{\okmark}{{\textbf{\textcolor[rgb]{0.1, 0.5, 0.1}{$\checkmark$}}}}
\renewcommand{\ngmark}{{\textbf{\color{red}{\ding{55}}}}}

\title{Extrapolating and Decoupling Image-to-Video Generation Models: \\ Motion Modeling is Easier Than You Think}

\renewcommand*{\thefootnote}{\fnsymbol{footnote}}

\author{
    Jie Tian\textsuperscript{1}\textsuperscript{\dag}\textsuperscript{*}
    \quad Xiaoye Qu\textsuperscript{1}\textsuperscript{\dag}
    \quad Zhenyi Lu\textsuperscript{1}\textsuperscript{\dag}
    \quad Wei Wei\textsuperscript{1}\textsuperscript{\Letter}
    \quad Sichen Liu\textsuperscript{1}
    \quad Yu Cheng\textsuperscript{2} \\
    \textsuperscript{1}School of Computer Science \& Technology, Huazhong University of Science and Technology, \\
    \textsuperscript{2}The Chinese University of Hong Kong \\
}
\vspace{-5mm}

\newcommand\blfootnote[1]{%
\begingroup
\renewcommand\thefootnote{}\footnote{#1}%
\addtocounter{footnote}{-1}%
\endgroup
}
\renewcommand*{\thefootnote}{\arabic{footnote}}
\begin{document}
\maketitle

\blfootnote{\dag\ Equal contribution; *\ Work done during an internship at Shanghai AI Laboratory; \Letter\  corresponding authors. }

\input{0_abstract}  
\input{1_intro}

\input{2_related}

\input{3_method}

\input{4_experiment}
\input{5_conclusion}

{
    \small
    \bibliographystyle{ieeenat_fullname}
    \bibliography{main}
}

\input{X_suppl}

\end{document}

%% file: preamble.tex
\usepackage{xcolor}
\usepackage{multirow}

\usepackage{booktabs}
\usepackage{pifont} 
\newcommand{\okmark}{{\textbf{\textcolor[rgb]{0.1, 0.5, 0.1}{$\checkmark$}}}}
\newcommand{\ngmark}{{\textbf{\textcolor{red}{\ding{55}}}}}

\usepackage{times}
\usepackage{latexsym}
\usepackage{amsfonts}
\usepackage{multirow}
\usepackage{array}
\usepackage{caption}
\usepackage{booktabs}
\usepackage{colortbl}
\usepackage{xcolor}
\usepackage{multicol}
\usepackage{bm}
\usepackage{graphicx}
\usepackage{amsthm,amsmath,amssymb}
\usepackage{mathrsfs}
\usepackage{enumitem}
\usepackage{marvosym}

%% file: 0_abstract.tex
\begin{abstract}

Image-to-Video (I2V) generation aims to
synthesize a video clip according to a given image and condition (e.g., text). 
The key challenge of this task lies
in simultaneously generating natural motions while preserving the original appearance of the images.
However, current I2V diffusion models (I2V-DMs) often produce videos with limited motion degrees or exhibit uncontrollable motion that conflicts with the textual condition. 
To address these limitations, we propose a novel Extrapolating and Decoupling framework, 
which introduces model merging techniques to the I2V domain for the first time.
Specifically, our framework consists of three separate stages:
(1) Starting with a base I2V-DM, we explicitly inject the textual condition into the temporal module using a lightweight, learnable adapter and fine-tune the integrated model to improve motion controllability.  
(2) We introduce a training-free extrapolation strategy to amplify the dynamic range of the motion, effectively reversing the fine-tuning process to enhance the motion degree significantly.
(3) With the above two-stage models excelling in motion controllability and degree, we decouple the relevant parameters associated with each type of motion ability and inject them into the base I2V-DM. 
Since the I2V-DM handles different levels of motion controllability and dynamics at various denoising time steps, we adjust the motion-aware parameters accordingly over time. 
Extensive qualitative and quantitative experiments have been conducted to demonstrate the superiority of our framework over existing methods. Code is available at \url{https://github.com/Chuge0335/EDG}

\end{abstract}

%% file: 1_intro.tex
\begin{figure}[t]
  \centering
  \includegraphics[width=0.9\linewidth]{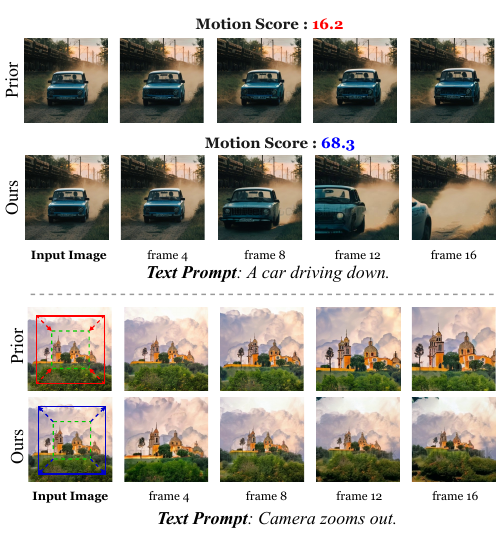}
   \caption{ 
In the first example, ConsistI2V \cite{ren2024consisti2v} yields a video with limited motion. In the second, DynamiCrafter \cite{xing2025dynamicrafter} fails to follow the text condition and lacks controllability. Our method, however, achieves improved motion degree and better controllability.
   }
   \label{fig:intro}
   \vspace{-5mm}
\end{figure}

\section{Introduction}

Recently, Image-to-Video generation (I2V) \cite{xing2025dynamicrafter,zhao2024identifying,ni2023conditional,zhang2024pia,guo2023animatediff} has attracted increasing attention. 
Given a single image and a text condition, 
it aims to convert the still image into videos that conform to text instruction while preserving the original appearance of the images. 
This paper focuses on open-domain images and supports free text as guidance, rather than limited class labels. 
This conditioned I2V task is substantially challenging as it struggles with simultaneously maintaining appearance consistency and producing natural motions in the generated frames.

Unlike image synthesis,  motion plays a significantly crucial role in image-to-video generation.
However, most existing works lack effective motion modeling when generating videos.
For instance, ConsistI2V \cite{ren2024consisti2v} enhances the first frame conditioning through spatiotemporal attention, leading to limited expression ability for modeling motion degree as the model tends to over-rely on the conditional image.
DynamiCrafter \cite{xing2025dynamicrafter} interacts the text embedding and conditional image embedding with the U-Net spatial layers through a dual cross-attention, neglecting the temporal property of the motion and suffering from motion controllability.  
In this way, the generated videos either contradict the motion described in the text or are still image sequences with slow change. As shown in Figure \ref{fig:intro}, the videos generated by the previous model show cars that barely move, or the videos of the castle contradict the textual descriptions.

To address the aforementioned issues, we present a novel Extrapolating and Decoupling framework to effectively model motion for image-to-video generation, improving both motion degree and motion controllability.
Specifically, it consists of three stages. 
Firstly, we improve motion controllability by explicitly injecting the textual condition into the I2V diffusion model (I2V-DM). A lightweight learnable adapter is introduced to enhance the temporal attention module.
Fine-tuning the integrated model improves motion controllability without altering the original I2V-DM architecture.
Secondly, we observe degraded motion dynamics in our fine-tuned model, a phenomenon has also been observed by \citet{zhao2024identifying}. 
To boost the motion degree without extra training, we introduce a novel training-free extrapolation strategy to restore and enhance motion degree dynamics by reversing the fine-tuning process.
Thirdly, we decouple the parameters contributing to motion controllability and dynamics from the previous two stages, selectively integrating them into the final model.
Recognizing that different denoising steps focus on varying motion ability and temporal consistency, we dynamically adjust the motion-aware parameters over time. 
With these three stages, we can generate videos with both natural motion and temporal consistency.

To verify the effectiveness of our framework, we perform experiments on the VBench I2V benchmark and UCF101 dataset. 
Extensive qualitative and quantitative experiments demonstrate the notable superiority of our framework over existing I2V methods. 
Furthermore, we offer discussion and analysis of some insightful designs in our framework. To sum up, our contributions are summarized as follows:

\begin{itemize}

\item We propose a novel Extrapolating and Decoupling framework for video generation, which models motion controllability and degree dynamics. To best of our knowledge, this is the first use of model merging techniques in the I2V domain.
\item We introduce a lightweight adapter to enhance motion controllability while preserving the base I2V-DM structure and propose a novel training-free extrapolation strategy to amplify motion degree. 
\item We make the first attempt to decouple parameters related to motion controllability and degree, dynamically injecting them into the I2V-DM across varying denoising steps.
\item We conduct extensive experiments on multiple
benchmarks, where our proposed framework consistently outperforms previous state-of-the-art methods. 

\end{itemize}

%% file: 2_related.tex
\section{Related Work}

\begin{figure*}[t]
    \centering
    \includegraphics[width=0.95\linewidth]{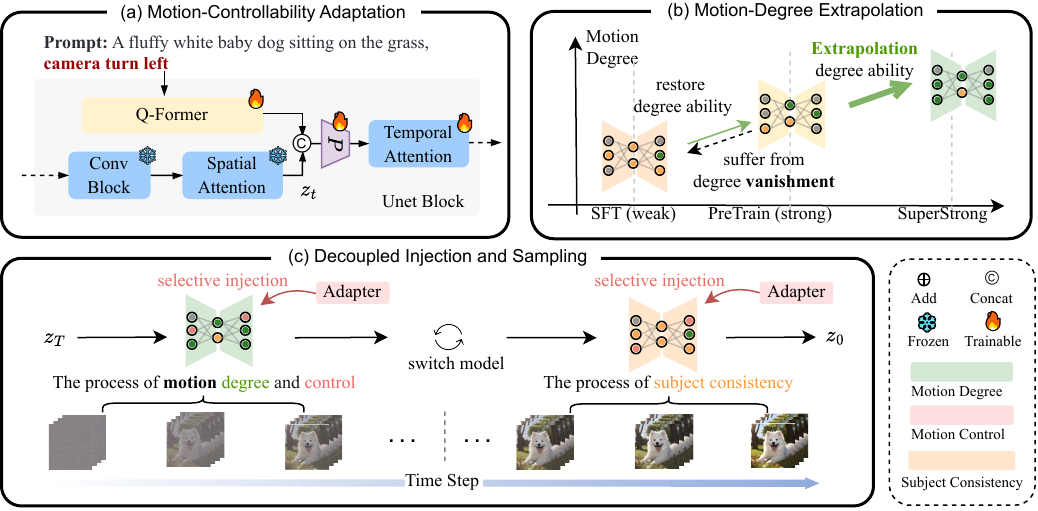} %
    \caption{ The overview of our framework, consisting of three stages.
    (a) We improve the motion controllability by injecting the textual conditions into the temporal attention module in I2V-DM model and then fine-tune the integrated model. 
    (b) The I2V-DM model suffers from a decrease in motion degree after fine-tuning. To mitigate it, 
    we propose a training-free extrapolation to boost the motion degree by reversing the fine-tuning progress.  
    (c) We decouple the relevant parameters contributing to motion controllability and motion degree from the last two-stage models. Moreover, we selectively inject these parameters into the I2V model along with the de-noising process. 
    }
    \label{fig:main framework}
    \vspace{-3mm}
\end{figure*}

\subsection{Motion Modeling in I2V Generation}

Diffusion models (DMs) \cite{ho2020denoising,sohl2015deep} have recently demonstrated remarkable generative capabilities in image and video generation \cite{he2023scalecrafter,nichol2021glide,ramesh2022hierarchical,rombach2022high,saharia2022photorealistic}.
In this paper, we focus on the image-conditioned video (I2V) generation task,
where motion modeling aims to predict object movements. Traditionally, optical flow estimates the displacement field between consecutive frames. To effectively model motion in I2V tasks, LFDM \cite{ni2023conditional} proposes a latent flow diffusion model that synthesizes a latent optical flow sequence to warp the given image in latent space for generating new videos. Motion-I2V \cite{shi2024motion} factorizes I2V into two stages: a diffusion-based motion field predictor for pixel trajectories and motion-augmented temporal attention to enhance 1-D temporal attention in video latent diffusion models. Recent works \cite{wang2024videocomposer,xing2025dynamicrafter,zhang2023i2vgen} enable diffusion models to handle motion modeling and video generation simultaneously. For instance, Dynamicrafter \cite{xing2025dynamicrafter} uses a dual-stream image injection mechanism to leverage the motion prior of text-to-video diffusion models. In this paper, we build on DynamiCrafter to design an Extrapolating and Decoupling framework to improve both motion controllability and degree.

\subsection{Model Merging}

Model merging, widely explored in natural language processing and image classification, typically combines multiple fine-tuned task-specific models into a unified multi-task model without additional training. 
Early methods \cite{draxler2018essentially,garipov2018loss,frankle2020linear,2022MergingModelsFisherWeighted,2023DatalessKnowledgeFusion,ainsworth2023git} leverage interpolation techniques to integrate multi-task knowledge effectively. 
Task-Arithmetic \cite{ilharco2022editing} extends these approaches by enabling complex arithmetic operations in weight space, offering finer control over the merging process. 
Recent studies \cite{ortiz-jimenez2023task,yadav2024ties,yang2024adamerging,yu2024language,lu2024twin} have further reduced interference during merging.
Beyond interpolation-based methods, recent works have shown that sparsifying \cite{yu2024language} or amplifying delta weights \cite{zheng2024weaktostrongextrapolationexpeditesalignment} can enhance model knowledge in merging. 
Additionally, merging techniques have been introduced to multimodal understanding such as question-answering \cite{shukor2023univalunifiedmodelimage,cheng2024damdynamicadaptermerging}. However, the video generation domain has remained unexplored until now.

\noindent We pioneer the introduction of model merging techniques into the video generation domain. Specifically, we leverage model merging to extrapolate motion degree knowledge and extract fine-grained decoupled knowledge, which are selectively merged at different diffusion stages.

%% file: 3_method.tex
\section{Method}

Given a reference image \( I_0 \) and a text prompt \( c \), image-to-video generation (I2V) targets at generating a sequence of subsequent video frames $V=$\( \{ \hat{I}_1, \hat{I}_2, \ldots, \hat{I}_N \} \). 
The key objective of I2V generation is to ensure that the generated video not only exhibits plausible motion but also faithfully preserves the visual appearance of the provided reference image. 
The overall framework of our method, shown in Figure \ref{fig:main framework}, consists of three stages. First, we improve motion controllability guided by the text condition. Second, we propose a training-free extrapolation strategy to enhance motion dynamics. Finally, we decouple parameters related to motion controllability and dynamics, allowing flexible injection during the denoising process to generate videos with natural motion and temporal consistency.

\subsection{Preliminary: Latent Video Diffusion Models}

Before introducing our framework, we first review the Latent Diffusion Models (LDMs) \cite{rombach2022high}.
Specifically, it extends diffusion models \cite{ho2020denoising} by operating in a more compact latent space rather than directly on pixels, reducing computational complexity. 
In LDMs, an input \( x_0 \) is first encoded into a latent representation \( z_0 \) using a pretrained Variational Autoencoder (VAE). This latent variable \( z_0 \) is then corrupted gradually over \( T \) steps by adding Gaussian noise, producing a sequence of noisy latents \( \{z_t\}_{t=1}^T \). The noise process follows \( z_t \mid z_0 \sim \mathcal{N}(z_t; \sqrt{\bar{\alpha}_t} z_0, (1 - \bar{\alpha}_t) I) \), where \( \bar{\alpha}_t = \prod_{i=1}^t \alpha_i \) controls the noise schedule. 
To recover the original latent, LDMs learn a reverse process that denoises \( z_t \) back to \( z_0 \) through a sequence of steps, guided by the objective:
\begin{equation}
\mathcal{L} = \mathbb{E}_{z_0, t, \epsilon} \left\Vert \epsilon_{\theta}(z_t, t) - \epsilon \right\Vert^2,
\end{equation}
where \( \epsilon \sim \mathcal{N}(0, I) \) represents Gaussian noise, \( t \) is a randomly sampled timestep, and \( \epsilon_{\theta}(z_t, t) \) is the learned noise estimator parameterized by \( \theta \).

In this paper, we adopt an open-source video LDM, DynamiCrafter \cite{xing2025dynamicrafter} as the base model.
Given a video \(x \in \mathbb{R}^{L \times 3 \times H \times W}\), we first encode it into a latent representation \(z = \mathcal{E}(x)\), where \(z \in \mathbb{R}^{L \times C \times h \times w}\) is derived on a frame-by-frame basis. Both the forward diffusion process \(z_t = p(z_0, t)\) and the backward denoising process \(z_t = p_{\theta}(z_{t-1}, c, t)\) are conducted in this latent space, with \(c\) representing potential denoising conditions such as text prompts. Ultimately, the generated videos are obtained through the decoder \(\hat{x} = \mathcal{D}(z)\).

\subsection{Motion-Controllability Adaptation}\label{sec:control}

In I2V generation, motion controllability is important for following textual conditions.
To model motion, current I2V-DMs \cite{xing2025dynamicrafter} typically employ spatial attention to share information within a single frame and temporal attention to capture time sequence consistency \cite{he2023latentvideodiffusionmodels}.
However, methods like DynamiCrafter \cite{xing2025dynamicrafter} only use text condition embedding in the spatial attention module, 
thus it does not share conditional information across different time sequences, limiting its motion controllability.

To alleviate this problem, we explicitly inject textual conditions into DynamiCrafter's temporal attention with a lightweight adapter, as depicted in Figure \ref{fig:main framework}(a). The spatial attention module remains unmodified as it already incorporates textual information. 
The adapter is non-intrusive and preserves the original model structure.
Specifically, we extracts spatial features $ z_t \in \mathbb{R}^{f \times d \times w \times h} $ and the text embeddings $ t_{\text{embd}} \in \mathbb{R}^{l \times d} $ from the spatial attention layer and the CLIP text encoder respectively, where $ f $ is the number of frames, $ d $ is the feature dimension, and $ w $, $ h $ are spatial dimensions. 
Then, the condition information in text embeddings are further compressed by a trainable Q-Former \cite{li2023blip} by retaining the motion-relevant tokens:
\begin{equation}
t'_{\text{embd}} = \text{Q-Former}(t_{\text{embd}}) \in \mathbb{R}^{f \times d}
\end{equation}
$t'_{\text{embd}}$ is then concatenated with spatial information $z_t$ along the feature dimension, projected through $W_{\text{proj}}$ to reduce the dimension to $d$, and passed through temporal attention to inject the text control information into each frame. 
We finetune the model on WebVid-2M \cite{Bain21}, updating only the adapter (the Q-Former and the linear projection $W_{\text{proj}}$) and temporal attention weights, ensuring minimal added parameters and efficient training.

\subsection{Motion-Degree Extrapolation}\label{sec:ex}

After fine-tuning I2V-DM DynamiCrafter with motion-controllability adaptation, we observe a reduction in motion degree, a common issue attributed to conditional image leakage \citep{zhao2024identifying}. During fine-tuning, Dynamicrafter concatenates the first frame with noise, which may cause the model to over-rely on the conditional image and neglect noisy inputs, especially at early timesteps. 
While adding time-dependent noise partially mitigates motion loss \citep{zhao2024identifying}, it requires retraining and has limited effectiveness.

To improve the motion dynamics beyond the pre-trained model, we introduce a novel \textit{training-free} and \textit{model-agnostic} extrapolation approach, as shown in Figure \ref{fig:main framework} (b), which extremely pushes the ``unlearning'' process of motion controllability. 
In this way, we can achieve a model with amplified motion degree dynamics.
Specifically, we estimate a degree-enhanced model, $\theta_{dyn}$, by extrapolating between a pre-trained model $\theta_{pre}$ and its fine-tuned version $\theta_\mathit{sft}$ with motion-controllability adaptation, amplifying their motion-dynamic differences:
\begin{equation}
    \theta_{dyn} = \theta_{pre} + \alpha (\theta_{pre} - \theta_\mathit{sft})
    \label{eq:extrapolate}
\end{equation}
Where the hyperparameter $\alpha$ controls the strength of this extrapolation. 
In this equation, we compute the difference of the model parameters except for the adapter introduced in Section \ref{sec:control}. 
This process requires only two checkpoints and obtaining the motion degree-enhanced model without training. 
To help further understand our extrapolating method, we provide a theoretical analysis below. 
 
\paragraph{{Theoretical Analysis.}} 
Assume that we have a motion degree metric score denoted as $\mathcal{D}$. 
Diffusion models are commonly regularized by a KL constraint in Gaussian distribution, implying that ${\theta_{pre}}$ lies within a small vicinity of ${\theta_\mathit{sft}}$. Therefore, a first-order Taylor expansion can be applied:

\begin{equation}
\begin{aligned}
    \mathcal{D}(\theta_{dyn}) &= \mathcal{D}(\theta_{pre} + \alpha \Delta \theta)\\
    &\approx \mathcal{D}(\theta_{pre}) + \alpha \left<\nabla \mathcal{D}(\theta_{pre}), \Delta \theta\right>\\
\end{aligned}
    \label{eq:taylor}
\end{equation}
where $\Delta$ denotes the parameter update from $\theta_\mathit{sft}$ to $\theta_{pre}$. The change in the motion degree is then given by:
\begin{equation}
\begin{aligned}
    \mathcal{D}(\theta_{dyn}) - \mathcal{D}(\theta_{pre}) &= \alpha \left<\nabla \mathcal{D}(\theta_{pre}), \Delta \theta\right>\\
    &\approx \alpha \left<\nabla \mathcal{D}(\theta_{pre}), \nabla \mathcal{D}(\theta_{pre})\right>\\
\end{aligned}
\end{equation}

Assuming that the parameter update \(\Delta\theta\) during fine-tuning is approximately aligned with the gradient of \(\mathcal{D}\) \cite{ho2022classifier,ilharco2022editing}, we can express \(\Delta\theta\) as:

\begin{equation} \Delta \theta \approx \gamma \nabla_\theta \mathcal{D}(\theta_{pre})
\end{equation}
Thus by substitution, we have:
\begin{equation}
\begin{aligned}
\mathcal{D}(\theta_{{dyn}}) - \mathcal{D}(\theta_{{pre}}) &\approx \alpha \left\langle \nabla_\theta \mathcal{D}(\theta_{{pre}}), \gamma \nabla_\theta \mathcal{D}(\theta_{{pre}}) \right\rangle \\
&= \alpha\gamma \left\langle \nabla_\theta \mathcal{D}(\theta_{{pre}}), \nabla_\theta \mathcal{D}(\theta_{{pre}}) \right\rangle \\
&= \alpha\gamma \left\Vert \nabla_\theta \mathcal{D}(\theta_{{pre}}) \right\Vert^2 \ge 0.
\end{aligned}
\end{equation}

Since the norm of the gradient \(\left\Vert \nabla_\theta \mathcal{D}(\theta_{{pre}}) \right\Vert^2\) is always non-negative, the dynamic motion degree metric \(\mathcal{D}\) shall increases after the extrapolation, provided that the pretrained weight \(\theta_{pre}\) is not at a local maximum of \(\mathcal{D}\).

\subsection{Decoupled Injection and Sampling}

While \textit{motion-controllability adaptation} and \textit{training-free degree extrapolation} can enhance motion control and diversity respectively, directly combining them poses a significant challenge due to their opposing optimization directions: fine-tuning for control versus reversed extrapolation for degree. 
To balance motion controllability and degree without compromising image consistency, we propose a decoupling strategy that separates degree-related and control-related parameters, applying them at distinct denoising stages.

\paragraph{Isolating Parameter Sets.}

Here, we categorize the obtained knowledge into three parameter sets that support core abilities: motion control ($\theta_{adt}$), motion degree ($\theta_{deg}$), and video consistency ($\theta_{con}$).
We aim to {integrate degree and control knowledge in the early denoising stage for long-term temporal planning, while emphasize consistency and control later for refining subject appearance.} However, they are embedded within $\theta_{pre}$ and $\theta_{sft}$.
Therefore, we need parameter isolation to decouple three types of knowledge.

We isolate these sets based on the following insight: 
(1) Based on the implicit gradient assumption (Section \ref{sec:ex}), $\theta_\mathit{sft} - \theta_{pre}$ captures the implicit control knowledge learned during the adaptation process, serves as $\theta_{adt}$ paired with (\(\theta_{adapter}\)), incorporating elements like Q-Formers and projectors. 
(2) {$\theta_\mathit{sft}$ exhibits degraded degree knowledge (11.67 \vs 68.54 for $\theta_{pre}$ in Table \ref{tab:vbench}) and primarily contains control and consistency knowledge. Therefore, consistency knowledge $\theta_{con}$ can be directly derived as $\theta_\mathit{sft} - \theta_{adt}$.
(3) {The extrapolated $\theta_{dyn}$ has similar control/consistency proportions as $\theta_{pre}$, therefore $\theta_{deg} = \theta_{dyn} - \theta_{pre}$ can remove control/consistency components, isolating degree knowledge, even though it originates from $\theta_\mathit{sft}$.}
}
(4) Fine-tuned parameters often contain redundant elements, which can cause conflicts during merging \cite{lu2024twin,yu2024language}. To minimize overlap between different knowledge while preserving performance,  
we apply DARE pruning \cite{yu2024language} at a pruning rate \(p\) (\textit{e.g.}, \(p = 70\%\)).
Distinct parameter groups are created using random masks $\mathcal M$ ampled from a Bernoulli distribution:
\begin{equation}
    \begin{aligned}
    \mathcal M_1 &\sim \text{Bernoulli(\textit{p})}, \quad \mathcal M_2 \sim \text{Bernoulli(\textit{p})}, \\
    \theta_{adt} &= \{ \theta_{adapter}\} \cup \{ \text{DARE}_{\mathcal M_2} (\theta_\mathit{sft} - \theta_{pre}) \}, \\
    \theta_{deg} &= \text{DARE}_{\mathcal M_1}(\theta_{dyn} - \theta_{pre}), \\
    \theta_{con} &= \theta_\mathit{sft} - \theta_{adt},
    \end{aligned}
    \label{eq:ips}
\end{equation}
Finally, Task-Arithmetic \cite{ilharco2022editing} is employed to inject control parameters \(\theta_{adt}\) into both \(\theta_{deg}\) and \(\theta_{con}\), deriving two enhanced models: the degree-enhanced model \(\theta_{dyn}^*\) and the consistency-enhanced model \(\theta_{con}^*\). :
\begin{equation}
    \begin{aligned}
        \theta_{dyn}^* &= \text{Task-Arithmetic}(\theta_{deg}, \theta_{adt}), \\
        \theta_{con}^* &= \text{Task-Arithmetic}(\theta_{con}, \theta_{adt}).
    \end{aligned}
\end{equation}


\paragraph{Decoupled Video Sampling.} 

During the denoising process for video generation, we typically observe that in the early stages, I2V models prioritize long-term temporal planning, including the motion of both subjects and textures, while later stages often focus more on refining subject appearance, which is mainly related to content consistency and overall image quality \cite{wu2024customcrafter, voynov2023pextendedtextualconditioning}. Therefore, in the first 
$K$ steps of the denoising process, we adjust the video generator to make it more dynamic-enhanced and control-enhanced to increase motion controllability. 
In the later stages of the denoising process, we re-weight the motion-related modules, allowing the model to focus more on restoring the specific details of each frame of the subject, thereby generating high-consistency videos.
Specifically, we define a generated model $\theta_{gen}$ by blending the degree-enhanced model $\theta_{dyn}^*$ and the consistency-merged model $\theta_{con}^*$ based on the denoising step $t$ from $T$ to $0$:

\begin{equation}
    \begin{aligned}
    \theta_{gen} &= \alpha_t \theta_{dyn}^* + (1 - \alpha_{t}) \theta_{con}^* , \\
    \alpha_t &= 
\begin{cases} 
1 & \text{if } t > T - K \\
0 & \text{otherwise}
\end{cases}
\end{aligned}
\end{equation}
Here, $T$ is the total number of denoising steps, and $K$ is a threshold that determines the switching point.

%% file: 4_experiment.tex
\begin{table*}[ht]
\caption{Quantitative results on the VBench I2V benchmark for various I2V-DM pretrained baselines and finetuning/inference techniques, including our methods based on DynamiCrafter. ``PT.'' in the table denotes whether the method requires video pretraining, and ``FT.'' denotes whether the method requires supervised finetuning on WebVid-2M. The global best and DynamiCrafter-based best performing metrics are highlighted in \textbf{bold} and \underline{underline}, respectively.}
    \centering
\resizebox{1.0\linewidth}{!}{%
    \begin{tabular}{@{}lcccccccc@{}}
    \toprule
    \textbf{Model} & \textbf{Avg.} & \textbf{PT.} & \textbf{FT.} & \textbf{I2V Sub.} & \textbf{I2V Back.} & \textbf{Video Quality} & \textbf{Motion Degree} & \textbf{Motion Control} \\ \midrule
    \textbf{ConsistI2V} (\citeauthor{ren2024consisti2v}) & 75.88 & \okmark & \ngmark & 95.82 & 95.95 & 85.74 & 18.62 & 33.92 \\
    \textbf{SEINE} (\citeauthor{chen2023seine}) & 76.56 & \okmark & \ngmark & 96.57 & 96.80 & 85.71 & 34.31 & 23.67 \\
    \textbf{Animate-Anything} (\citeauthor{dai2023animateanythingfinegrainedopendomain}) & 74.61 & \okmark & \ngmark & \textbf{98.76} & {98.58} & \textbf{88.84} & 2.68 & 13.08 \\
    \textbf{VideoCrafter} (\citeauthor{chen2023videocrafter1opendiffusionmodels}) & 76.96 & \okmark & \ngmark & 91.17 & 96.94 & 87.55 & 22.60 & 33.60 \\
    \textbf{SVD-XT-1.0} (\citeauthor{blattmann2023stable}) & - & \okmark & \ngmark & 97.52 & 97.63 & 86.54 & 52.36 & - \\
    \textbf{SVD-XT-1.1} (\citeauthor{blattmann2023stable}) & - & \okmark & \ngmark & 97.51 & 97.62 & 86.66 & 43.17 & - \\
    \textbf{I2VGen-XL} (\citeauthor{zhang2023i2vgen}) & 76.00 & \okmark & \ngmark & 96.48 & 96.83 & 87.02 & 26.10 & 18.48 \\
    \midrule
    \textbf{DynamiCrafter} (\citeauthor{xing2025dynamicrafter}) & 79.93 & \okmark & \ngmark & 94.91 & 95.30 & 84.95 & 68.54 & 30.85 \\
    \textbf{DynamiCrafter-Naive FT} & 72.91 & \ngmark & \okmark & \underline{97.92} & \underline{\textbf{98.70}} & 83.52 & 11.67 & 17.82 \\
    \textbf{DynamiCrafter-CIL Infer} (\citeauthor{zhao2024identifying}) & 80.47 & \ngmark & \ngmark & 96.96 & 93.34 & 85.28 & 69.43 & 31.29 \\
    \textbf{DynamiCrafter-CIL FT} (\citeauthor{zhao2024identifying}) & 77.55 & \ngmark & \okmark & 96.90 & 97.79 & \underline{87.10} & 25.69 & 30.50 \\
    \midrule
    \textbf{DynamiCrafter-Ours} & {84.25} & \ngmark & \okmark & 97.03 & 96.62 & 86.22 & {87.64} & \underline{\textbf{43.87}} \\ 
    \textbf{DynamiCrafter-Ours w/ CIL Infer}& \underline{\textbf{84.81}} & \ngmark & \okmark  & {97.11} & 95.45 & \underline{87.10} & \underline{\textbf{89.21}} & {43.61} \\
    \bottomrule
    \end{tabular}
}
\label{tab:vbench}
\end{table*}

\section{Experiment}

\subsection{Setup}

In all experiments, videos and images are resized to a \(320 \times 512\) center-cropped format. During model training, we set the learning rate to \({1 \times 10^{-5}}\), a batch size of 64, and trained for 100K steps on 8 A100 GPUs. For inference, we use DDIM \cite{song2022ddim} with 50 steps and a classifier-free guidance scale of 7.5, generating 16 frames for a 2-second video.

\noindent \textbf{Datasets.} We use the WebVid-2M dataset \cite{Bain21} for supervised fine-tuning, which contains two million video-text pairs from stock footage websites. For motion controllability annotation, we use CoTracker \cite{karaev2024cotrackerbettertrack} to annotate motion information on the WebVid-2M dataset.

\noindent \textbf{Evaluation.} The evaluation benchmarks include VBench \cite{ren2024consisti2v} and UCF101 \cite{soomro2012dataset}. VBench is a comprehensive benchmark for video generation, incorporating recent models and extensive evaluation metrics. UCF101 contains 101 human action categories with diverse background and lighting variations. 
For VBench, following previous work \cite{zhao2024identifying, liu2024mardini}, we utilize the official dataset to assess the model across several metrics: I2V-Subject Consistency (I2V Sub.), I2V-Background Consistency (I2V Back.), video quality, camera motion, and dynamic degree. The video quality evaluation considers dimensions including Subject Consistency, Background Consistency, Smoothness, Aesthetic Score, Imaging Quality, and Temporal Flickering. 
The Avg. score is calculated as $ \frac{1}{10}*(\text{I2V Sub.} + \text{I2V Back.} + 6* \text{Video Quality} + \text{Motion Degree} + \text{Motion Control})$, which is consistent with VBench I2V original settting. 
For UCF101, similar to \cite{xing2025dynamicrafter, wu2024customcrafter}, we use Fréchet Video Distance (FVD) \cite{unterthiner2019fvd} and Inception Score (IS) \cite{saito2017is} to assess video quality.

\begin{table}[h]
\caption{Quantitative evaluation of the DynamiCrafter-based method on the UCF-101 benchmark. \label{tab:ucf}}
\centering
    \resizebox{0.95\linewidth}{!}{%
\begin{tabular}{@{}lccc@{}}
\toprule
\textbf{Model} & \textbf{FVD}$\downarrow$ \ & \textbf{IS}$\uparrow$ & \textbf{Motion Degree}$\uparrow$ \\ \midrule 
\textbf{DynamiCrafter}           & \textbf{288.01}                  & 13.31                  & 68.17                            \\ 
\textbf{DynamiCrafter-Naive FT}  & 373.77                  & 13.19                  & 47.66                            \\
\textbf{DynamiCrafter-CIL FT}  & 409.69                  & 13.68                  & 53.78                            \\ \midrule
\textbf{DynamiCrafter-Ours} & 324.79                       &   \textbf{13.71}                      &        \textbf{77.93}                        \\ \bottomrule
\end{tabular}
}
\end{table}

\subsection{Quantitative Evaluation}

\paragraph{VBench Results Analysis.} 

We compare our method with two types of approaches on VBench. The first kind of comparing methods include those requiring large-scale video pretraining, such as Animate-Anything \cite{dai2023animateanythingfinegrainedopendomain}, SVD \cite{blattmann2023stable}, and DynamiCrafter \cite{xing2025dynamicrafter}. 
The other comparison methods are finetuning/inference strategies based on DynamiCrafter, including naive finetuning with the same training setup as our method (Naive FT) and Conditional Image Leakage (CIL) methods \cite{zhao2024identifying} (CIL Infer and CIL FT).
Notably, our method performs best in Motion Degree and Motion Control across all baselines, even including large-scale pretrained methods, as shown in Table \ref{tab:vbench}.
We observe that Naive FT causes a significant degradation in Motion Degree (11.67 vs 68.54), consistent with previous findings \cite{zhao2024identifying}. Naive FT achieves the highest I2V Subject and Background Consistency (I2V Sub. and I2V Back.) among the DynamiCrafter-based methods because lower motion degree results in higher consistency scores. Interestingly, motion control is also negatively affected in Naive FT (17.82 vs 11.54). We speculate that this is due to the insufficient motion degree, which hinders the model's ability to follow control instructions effectively.
The CIL method can mitigate these effects, either by employing the inference strategy on the pre-trained model (CIL Infer) or by repairing the Naive FT (CIL FT). However, its enhancements in Motion Control and Motion Degree are limited. In contrast, our method, combining adaptation and extrapolation, significantly improves both motion degree and motion control (11.67 vs 87.64) while preserving similar video quality (87.10). Furthermore, our method is orthogonal to the CIL Infer method, and integrating it with our method further boosts performance (Avg. 84.25 vs 84.81).

\begin{figure}[t]
  \centering
  \includegraphics[width=1.0\linewidth]{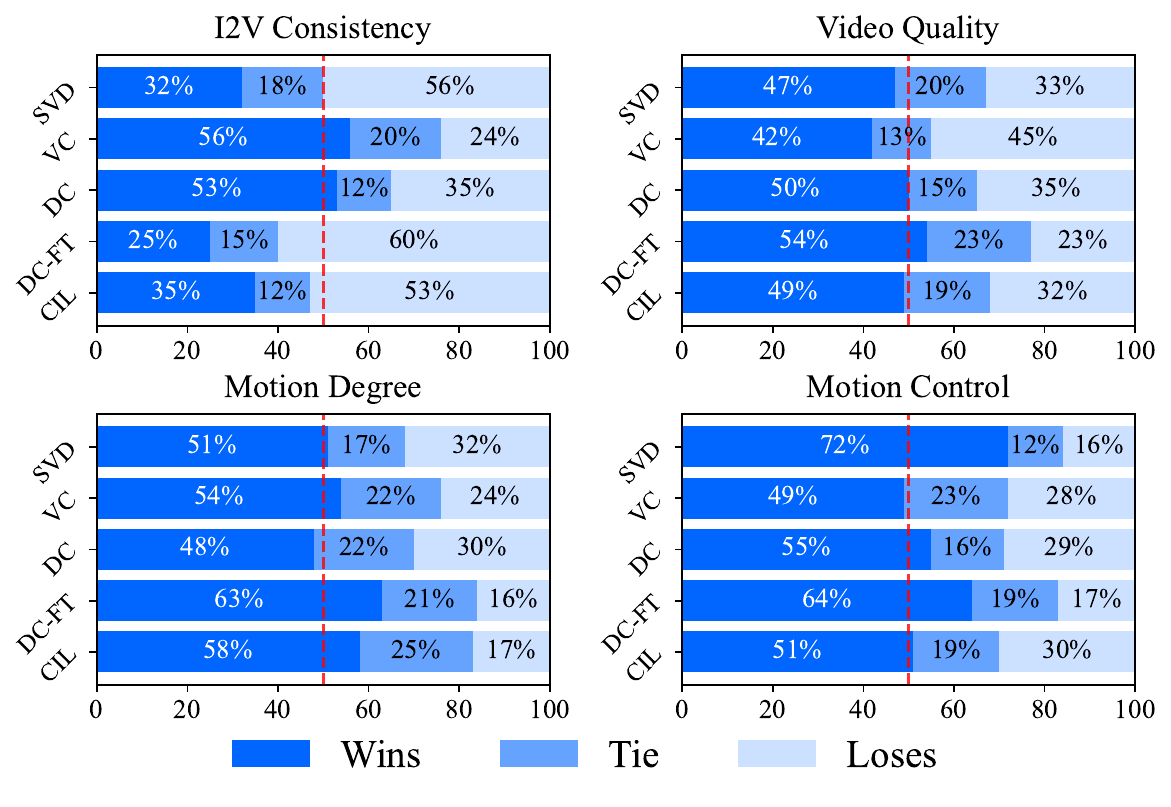}
   \caption{Human Evaluation of our method Compared to SVD, VideoCrafter (VC), DynamiCrafter (DC), and its variant of Native Fine-Tuned (DC-FT) and CIL-Based Methods.}
   \label{fig:onecol}
\end{figure}

\begin{figure*}[!h]
    \centering
    \begin{subfigure}[b]{0.87\linewidth}
        \centering
        \includegraphics[width=\linewidth]{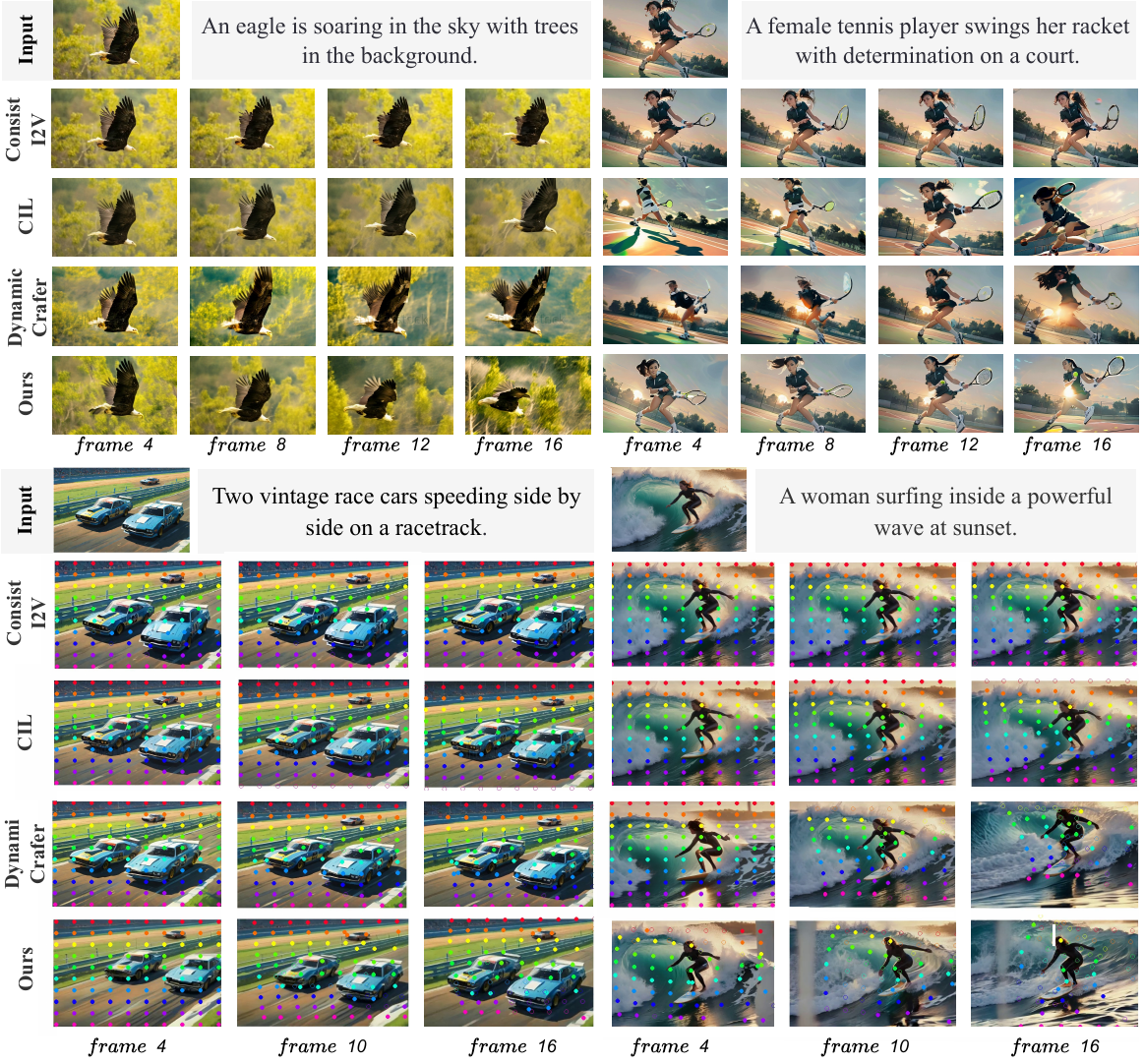}
        \label{fig:qualitative2}
    \end{subfigure}
    \caption{In the visual comparisons, we generated 16-frame videos using DynamiCrafter, CIL, Consistent-I2V, and our model. We selected 3 to 4 frames at regular intervals for display at the top. The upper section visualizes the overall aesthetic quality, while the lower section uses Optical Flow to illustrate the dynamic qualities of the generated videos for each method.}
    \label{fig:qualitative}
    \vspace{-2mm}
\end{figure*}

\paragraph{UCF101 Results Analysis.} In the UCF101 dataset, our method still exhibits superior motion degree ability despite the low resolution and poor image quality. Both Naive FT and CIL FT suffer from degradation in Fréchet Video Distance (FVD), as shown in Table \ref{tab:ucf}. Unlike the performance on VBench, CIL performs worse than Naive FT on UCF101, likely due to the added noise schedule disrupting the already noisy and low-resolution images. Our method, however, is more stable in FVD and outperforms in Inception Score (IS) and Motion Degree.

\paragraph{User Study.} As shown in Figure \ref{fig:onecol}, our method excels in Motion Degree and Motion Control while remaining competitive in video quality and I2V consistency. The results are consistent with the above findings, supporting that our strategies enhance both video motion dynamics and control while preserving quality and consistency, yielding superior overall performance.

\begin{table*}[t]
\caption{Ablation study of the three stages of our framework on VBench.}
    \centering
\resizebox{0.92\linewidth}{!}{%
    \begin{tabular}{@{}ccc|ccccccccc@{}}
    \toprule
    \textbf{(a) Adaptation} & \textbf{(b)Extrapolation } & \textbf{(c) Decouple} & \textbf{I2V Sub.} & \textbf{I2V Back.} & \textbf{Video Quality} & \textbf{Motion Degree} & \textbf{Motion Control} \\ \midrule
      \okmark & \ngmark & \ngmark & 97.51 & 96.39 & 86.61 & 22.85 & 49.31 \\
      \ngmark & \okmark & \ngmark & 94.14 & 95.62 & 83.52 & 98.21 & 17.63 \\
      \ngmark & \ngmark & \okmark & 94.60 & 96.21 & 83.95 & 69.61 & 32.37 \\ 
      \okmark & \okmark & \ngmark &91.11 & 92.86 & 84.32 & 71.68 & 26.71 \\ 
      \ngmark & \okmark & \okmark & 96.01 & 94.60 & 84.34 & 93.33 & 21.85 \\ 
      \okmark & \okmark & \okmark & 97.03 & 96.62 & 86.22 & 87.64 & 43.87 \\\bottomrule
    \end{tabular}
}
\label{tab:ab}
\end{table*}

\subsection{Qualitative Evaluation}

To empirically evaluate the visual quality of our method, we compare it with several state-of-the-art image-to-video generation approaches,
including DynamiCrafter \cite{xing2025dynamicrafter}, CIL \cite{zhao2024identifying}, and ConsistI2V \cite{ren2024consisti2v}. In Figure~\ref{fig:qualitative}, we observe that our method generates temporally coherent videos with a sufficient range of motion that closely adhere to the input text conditions. In contrast, DynamiCrafter exhibits inconsistencies in text control, likely due to challenges in capturing the semantic content of the input image. The ConsistI2V method produces videos that remain static over time, with the color points on the figures almost unchanged, indicating a lack of motion dynamics and control. While CIL generates videos that semantically resemble the input images and have enhanced motion compared to ConsistI2V, its improvements are limited. CIL does not sufficiently vary the content dynamically to match the text prompts, resulting in limited motion and control. Our method, however, preserves image consistency while achieving a large motion degree and exhibiting good controllability.

\subsection{Ablation Study}

We demonstrate the importance of our extrapolation and decoupling methods in Table \ref{tab:ab}:
(1) On top of our adaptation training, using our extrapolation and decoupling techniques significantly improves the motion degree, from 22.85 to 87.64.
(2) Using extrapolation alone can drastically improve the motion degree (69.61 vs 93.33), but it comes at the cost of degradation in I2V-Subject Consistency (I2V Sub.) and I2V-Background Consistency (I2V Back.), as well as motion control. This is because reversing the training process of the pre-trained model to the SFT model unlearns the camera motion ability, causing significant interference if directly merged \cite{ilharco2022editing}.
(3) Using the decoupling technique alone can improve the motion degree while preserving other performance metrics. This indicates that the I2V model indeed exhibits functional differentiation during the sampling process.
(4) Adding adaptation, though it increases training costs, greatly enhances the Motion Control aspect of the model. The degradation in motion degree indicates that direct supervision is suboptimal and can lead to minor motion dynamics, a finding consistent with previous research \cite{zhao2024identifying}.
(5) When paired with the decoupling technique, we achieve optimal results, integrating high motion degree and control ability while preserving consistency and video quality.

\begin{figure}[th]
  \centering
  \begin{subfigure}[b]{0.92\linewidth}
    \centering
    \includegraphics[width=\linewidth]{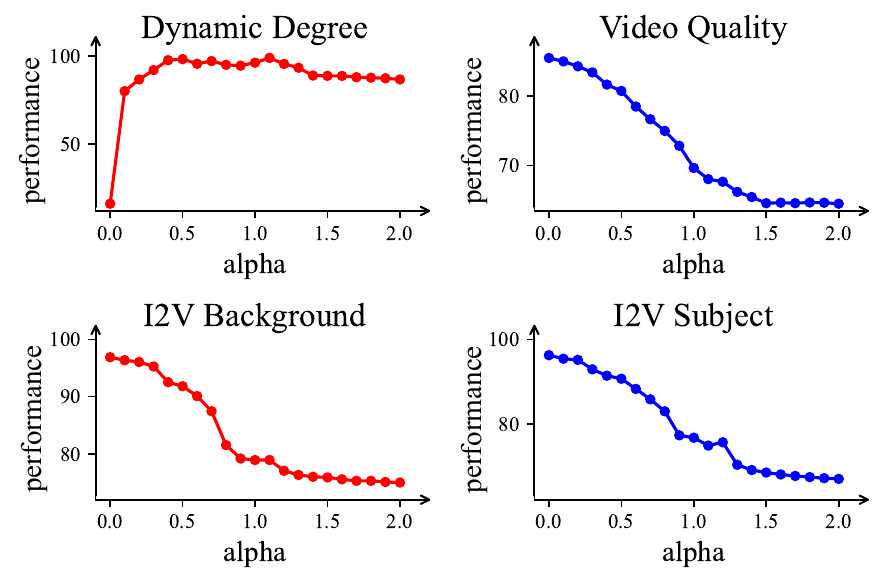}
    \caption{Effect of extrapolate strength $\alpha$.}
    \label{fig:alpha}
  \end{subfigure}
  \hfill
  \begin{subfigure}[b]{0.92\linewidth}
    \centering
    \includegraphics[width=\linewidth]{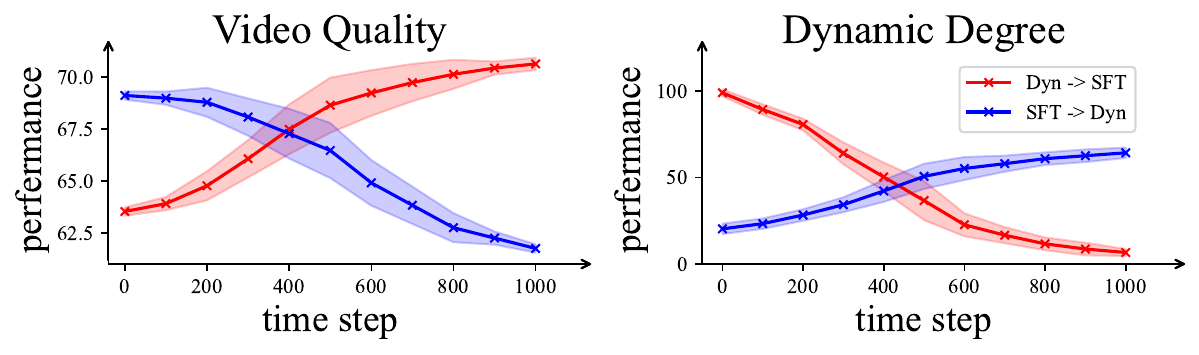}
    \caption{Effect of different switching time and model choices.}
    \label{fig:schedule}
  \end{subfigure}
  \caption{The impact of different extrapolate strengths and different sampling strategies.}
  \label{fig:combined}
  \vspace{-4mm}
\end{figure}

\subsection{Effect of Extrapolate Strength}

In our extrapolation experiments, we analyzed the impact of different \(\alpha\) values on performance using Equation \ref{eq:extrapolate}. As shown in Figure \ref{fig:alpha}, as \(\alpha\) increases, the motion degree first rapidly rises within the 0-0.7 range, peaks, and then gradually declines between 1.2-2.0. Simultaneously, quality and consistency metrics decline. This indicates that extrapolation enhances motion degree but sacrifices quality and consistency. This aligns with the findings in Section 4.4 and underscores the importance of appropriate decoupling.

\subsection{Analysis on Decoupled Sampling Method}

To study the impact of switch model time and choice, we conducted experiments illustrated in Figure \ref{fig:schedule}. We outline two model switch strategies during the denoising process from time \( T \) to 0: (1) Initially using \(\theta_{dyn}^*\) and then switching to \(\theta_{sft}^*\). (2) Starting with \(\theta_{sft}^*\) and then transitioning to \(\theta_{dyn}^*\).
The performance trends are shown in Figure \ref{fig:schedule}. The first strategy, which initially uses \(\theta_{dyn}^*\), results in higher motion degree with a later switch (smaller \( T \)), but with poorer quality. Conversely, an earlier switch (larger \( T \)) reduces the degree. The second strategy, introducing \(\theta_{dyn}^*\) later, has a less pronounced impact on the degree. This aligns with our hypothesis in Section 3.4, which suggests that the early sampling phase significantly influences the degree.

%% file: 5_conclusion.tex
\section{Conclusion}

In this paper, we introduce a novel Extrapolating and Decoupling framework for image-to-video generation. Our framework consists of three stages and can effectively model the motion control and motion degree. In the first stage, we explicitly inject the textual condition into I2V-DM to improve motion controllability. Then, we devise a training-free extrapolation strategy to boost the motion degree. Finally, we decouple the motion-relevant parameters and dynamically inject them into the I2V-DM along different de-noising steps. With our framework, we can generate videos with both natural motion and temporal consistency. To the best of our knowledge, we are the first to apply merging techniques in video generation.

%% file: X_suppl.tex
\clearpage
\setcounter{page}{1}
\maketitlesupplementary

\section{More Relative Research in Video Diffusion Models}

To extend Diffusion models (DMs) to video generation, the first video diffusion model (VDM) \cite{ho2022video} has been proposed, which utilizes a spacetime factorized U-Net to model low-resolution videos in pixel space. Imagen-Video \cite{ho2022imagen} introduces efficient cascaded DMs with v-prediction for producing high-definition videos. 
To mitigate training costs, subsequent research \cite{he2022latent,wang2023lavie,zhou2022magicvideo,blattmann2023align} has focused on transferring T2I techniques to text-to-video (T2V) \cite{ge2023preserve,luo2023videofusion,singer2022make,zhang2023show}, as well as on developing VDMs in latent or hybrid pixel-latent spaces.
Similar to the addition of controls in text-to-image (T2I) generation \cite{mou2024t2i,shi2024instantbooth,ye2023ip,zhang2023adding}, the introduction of control signals in text-to-video (T2V) generation, such as structure \cite{xing2024make,esser2023structure}, pose \cite{ma2024follow,zhang2023controlvideo} has garnered increasing attention. Nonetheless, visual image conditions in video diffusion models (VDMs) 
\cite{tang2024any,yin2023dragnuwa}, remain under-explored. Recent works, including Seer \cite{gu2023seer}, VideoComposer \cite{wang2024videocomposer}, have investigated image conditions for image-to-video synthesis. 
However, these approaches either focus on curated domains like indoor objects \cite{wang2024videocomposer} or struggle to produce temporally coherent frames and realistic motions, often failing to preserve visual details of the input image \cite{zhang2023i2vgen}. Recent proprietary T2V models \cite{molad2023dreamix,singer2022make,villegas2022phenaki,yu2023magvit,chen2024accelerating} show potential for extending image-to-video synthesis but often lack adherence to the input image and suffer from unrealistic temporal variations. In this paper, we focus on the image-conditioned video generation task.

\section{Dataset Selection}

We choose WebVid for fair comparison with prior work
CIL \cite{zhao2024identifying} and its wide adoption by DynamiCrafter \cite{xing2025dynamicrafter},
Consisti2v \cite{ren2024consisti2v}, and Motion-i2v \cite{shi2024motion}, due to its uniform resolution and sufficient size. Other datasets like Panda-70M \cite{chen2024panda70mcaptioning70mvideos} have watermarks and blurriness, OpenVid \cite{nan2024openvid1mlargescalehighqualitydataset} lacks resolution consistency, Vript \cite{yang2024vriptvideoworththousands} is too small for effective training.

\section{Generalization of Our Framework.} 

To demonstrate the generalization, we further perform experiments on SVD \cite{blattmann2023stable}. As shown in Table \ref{tab:svd}, our method also delivers performance improvements on motion degree and motion control.

\begin{table}[h]
\scriptsize
\caption{Applying our method in Stable-Video-Diffusion \label{tab:svd}}
    \centering
\resizebox{0.9\linewidth}{!}{%
    \begin{tabular}{@{}lcccccccc@{}}
    \toprule
    \textbf{Model} & \textbf{Video Quality} & \textbf{Motion Degree} & \textbf{Motion Control} \\ \midrule
    \textbf{SVD} & 66.38 & 41.94 & 20.41 \\
    \textbf{SVD-Ours} & \textbf{67.24} & \textbf{55.16} & \textbf{36.53} \\ 
    \bottomrule
    \end{tabular}
}
\vspace{-1mm}
\end{table}

\section{Model Merging Method}\label{app:iso}

\paragraph{DARE-Pruning \cite{yu2024language}} 

DARE employs a parameterized Bernoulli distribution to sample a sparse mask \(\boldsymbol{m}^t\), which is then applied to the parameters \(\boldsymbol{\delta}\) and rescaled by the mask rate \(p\):

\begin{equation}
\begin{gathered}
    \boldsymbol{m}^t \sim \text{Bernoulli}(p), \\
    \widetilde{\boldsymbol{\delta}}^t = \boldsymbol{m}^t \odot \boldsymbol{\delta}^t, \\
    \hat{\boldsymbol{\delta}}^t = \frac{\widetilde{\boldsymbol{\delta}}^t}{1 - p}.
\end{gathered}
\end{equation}

\paragraph{Task-Arithmetic \cite{ilharco2022editing}} 

Task-Arithmetic introduces the concept of “task vectors." A task vector is obtained by subtracting the weights of a pre-trained model from the weights of the same model after fine-tuning. Performance on multiple tasks can be improved by combining vectors from different tasks. Formally, let \(\theta_{\text{pre}} \in \mathbb{R}^d\) be the weights of a pre-trained model, and \(\theta_{\mathrm{ft}}^t \in \mathbb{R}^d\) the weights after fine-tuning on task \(t \in \{1, \ldots, T\}\). The task vector \(\tau_t \in \mathbb{R}^d\) is given by:

\begin{equation}
\tau_t = \theta_{\text{ft}}^t - \theta_{\text{pre}}
\end{equation}

We can obtain a multi-task version of the model \(\theta_m\) by summing the task vectors:

\begin{equation}
\theta_m = \theta_{pre} + w \sum_{t=1}^{T} \tau_t
\end{equation}
where $w$ is a hyperparameter.

\section{Details of User Study.}

\begin{figure}[t]
    \centering
    \begin{subfigure}[b]{0.91\linewidth}
        \centering
        \includegraphics[width=\linewidth]{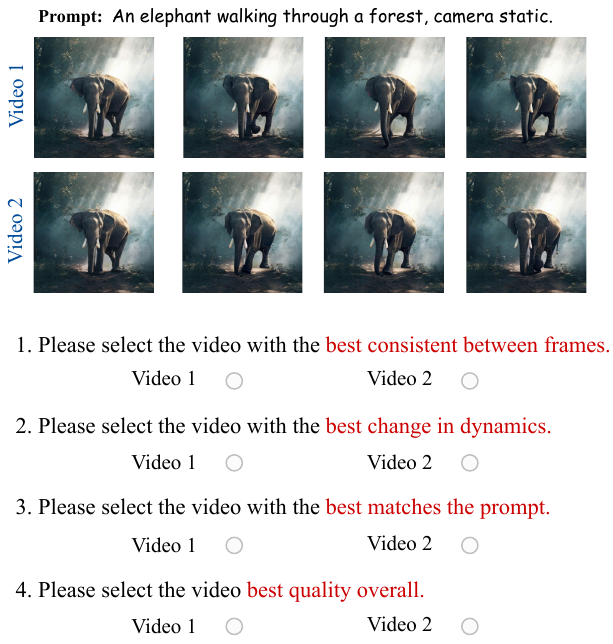}
        \label{fig:qualitative2}
    \end{subfigure}
    \caption{Example of user study questionnaires.}
    \label{fig:survey}
    \vspace{-1mm}
\end{figure}

For user study, we randomly select input image and prompt pairs in Vbench and then generate videos by using Ours, 
SVD \cite{blattmann2023stable} and VideoCrafter \cite{chen2023videocrafter1opendiffusionmodels} with DynamiCrafter\cite{xing2025dynamicrafter}, CIL \cite{zhao2024identifying} and DC-FT (fine-tuned DynamiCrafter).
In the setup, we conducted pairwise comparisons between our model and other methods, inviting users to evaluate and select the superior one in terms of quality, consistency, dynamism, and instruction following. 
We show an illustration of the question cases in Figure~\ref{fig:survey}. There are 40 video comparisons, comprising a total of 160 questions, with the order of the synthetic videos being shuffled. The survey involved a total of 200 participants.
Following the approach outlined in \cite{zhao2024identifying}, we calculated the preference rates, with the results presented in the main paper.

For camera motion, each command was executed 5 times on the 8 images, and the average success rate for each category was then computed. The human evaluation performance is shown in Table \ref{tab:camera_commands} and our method improves all categories over DynamiCrafter.
Notably, as the training data for ``Zoom out'' is extremely scarce in the original DynamiCrafter (0.26\%), both the DynamiCrafter and our method have relatively low scores on the zoom-out task and are not entirely flawless, although our method can improve it. 

\begin{table}[h]
\scriptsize
\caption{Camera Movement and Dataset Distribution.} 
\centering
\resizebox{1.0\linewidth}{!}{%
    \begin{tabular}{@{}ccccccc@{}}
    \toprule
    \multirow{2}{*}{\textbf{Model}} & \textbf{Pan left} & \textbf{Pan right} & \textbf{Tilt up} & \textbf{Tilt down} & \textbf{Zoom in} & \textbf{Zoom out} \\ 
    & (6.91\%) & (7.27\%) & (4.07\%) & (1.65\%) & (0.94\%) & (0.26\%) \\ \midrule
    DynamiCrafter & 0.43 & 0.53 & 0.53 & 0.28 & 0.33 & 0.08 \\
    Ours & \textbf{0.45} & \textbf{0.55} & \textbf{0.60} & \textbf{0.35} & \textbf{0.53} & \textbf{0.33} \\ \bottomrule
    \end{tabular}
}
\label{tab:camera_commands}
\vspace{-2mm}
\end{table}

\section{Examples of Different Scenario.}

\begin{figure}[t]
    \centering
    \begin{subfigure}[b]{0.91\linewidth}
        \centering
        \includegraphics[width=\linewidth]{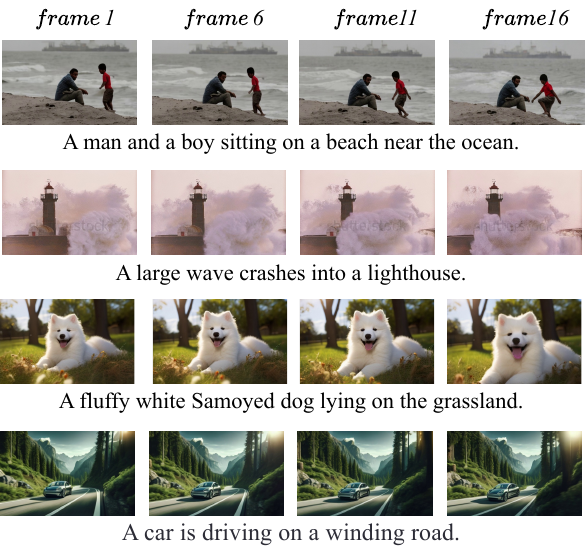}
        \label{fig:qualitative2}
    \end{subfigure}
    \caption{The figure above illustrates the performance of our model in generating videos across various categories, including humanities, natural phenomena, animals, and modern transportation.}
    \label{fig:sup_demo1}
    \vspace{-1mm}
\end{figure}
In this section, we present additional examples of generated videos, as shown in Figure~\ref{fig:sup_demo1}. The examples cover various real-world scenes, including human figures, natural phenomena, animals, and car movements. The human figures are generated with complete and fluid actions. In the second row, the generated waves submerge a lighthouse. In the third and fourth rows, the model preserves high consistency.

\begin{figure}[t]
    \centering
    \begin{subfigure}[b]{0.91\linewidth}
        \centering
        \includegraphics[width=\linewidth]{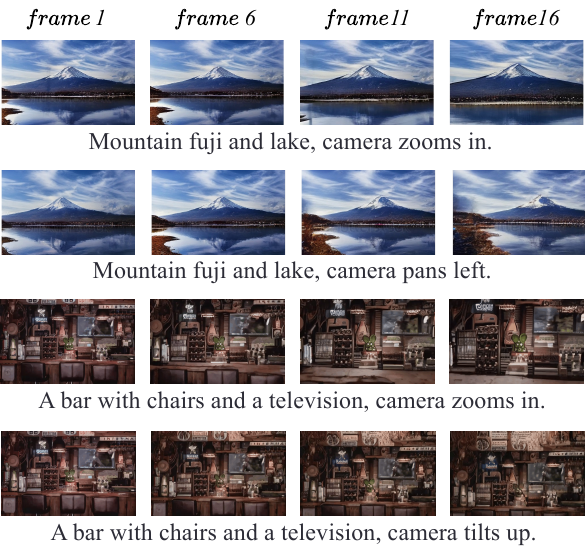}
        \label{fig:qualitative2}
    \end{subfigure}
    \caption{Visualization of Controllable Camera Movement with Different Text Prompts. Examples featuring simple and distinct main subjects are illustrated above, while those showcasing complex and disordered backgrounds are depicted below.}
    \label{fig:sup_demo2}
    \vspace{-1mm}
\end{figure}

For the same image input, different text instructions are used to control the variations in the actions within the generated videos. In the top two rows of Figure~\ref{fig:sup_demo2}, an image of Mount Fuji is fed into the model, with the text instructions “camera zooms in" and “camera pans left" appended to control the corresponding camera movements in the output. The bottom two rows provide examples of camera control in more complex scenes, utilizing similar instructions such as “camera zooms in" and “camera tilts up." Despite the high visual similarity, the generated videos respond well to these instructions.